\documentclass[letterpaper, 10 pt, conference]{ieeeconf}  
\usepackage[left=1.91cm,right=1.91cm,top=1.91cm,bottom=1.91cm]{geometry}
\IEEEoverridecommandlockouts                              
\usepackage{graphics} 
\usepackage{epsfig} 
\usepackage{mathptmx} 
\usepackage{times} 
\usepackage{amsmath} 
\usepackage{amssymb}  
\usepackage{xcolor}
\usepackage{multirow}
\usepackage{booktabs}
\usepackage[hidelinks]{hyperref}
\usepackage{cleveref}
\usepackage{comment}

\def\etal{\textit{et al.}}

\title{\LARGE \bf
F3 Hand: A Versatile Robot Hand Inspired by \\ Human Thumb and Index Fingers
}

\author{Naoki Fukaya$^{\dagger}$, Avinash Ummadisingu$^{\dagger}$, Guilherme Maeda$^{\dagger}$ and Shin-ichi Maeda$^{\dagger}$
	\thanks{$^{\dagger}$ Preferred Networks, Inc. 
	(As of the publication of this paper, Guilherme Maeda is affiliated with Sony AI.)
		{\tt\small \{fukaya, ummavi, ichi\}@preferred.jp, guilherme.maeda@sony.com}}
}

\begin{document}

\maketitle
\thispagestyle{empty}
\pagestyle{empty}

\begin{abstract}
It is challenging to grasp numerous objects with varying sizes and shapes with a single robot hand.
To address this, we propose a new robot hand called the ``F3 hand'' inspired by the complex movements of human index finger and thumb.
The F3 hand attempts to realize complex human-like grasping movements by combining a parallel motion finger and a rotational motion finger with an adaptive function.
In order to confirm the performance of our hand, we attached it to a mobile manipulator - the Toyota Human Support Robot (HSR) and conducted grasping experiments.
In our results, we show that it is able to grasp all YCB objects (82 in total), including washers with outer diameters as small as $6.4~\mathrm{mm}$.
We also built a system for intuitive operation with a 3D mouse and grasp an additional 24 objects, including small toothpicks and paper clips and large pitchers and cracker boxes.
The F3 hand is able to achieve a 98\% success rate in grasping even under imprecise control and positional offsets.
Furthermore, owing to the finger's adaptive function, we demonstrate characteristics of the F3 hand that facilitate the grasping of soft objects such as strawberries in a desirable posture.
\end{abstract}

\section{INTRODUCTION}
\begin{figure}[t]
	\centering
	\includegraphics[width=0.95\linewidth, draft=false]{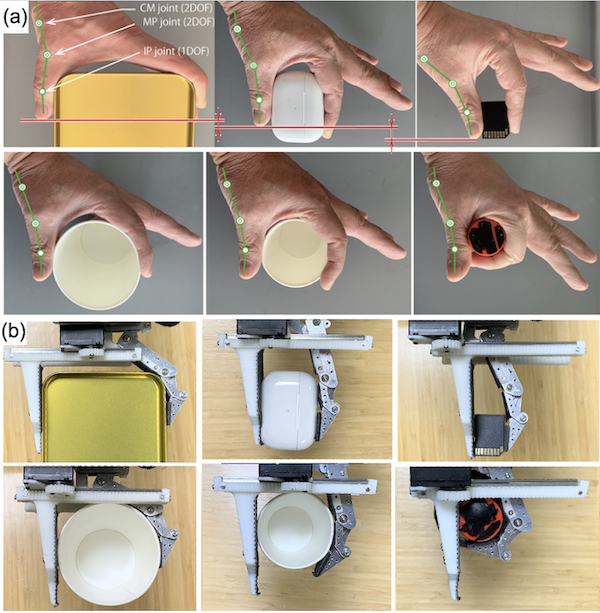}
	\caption{
	(a) The posture of the thumb when a human grasps an object.
    Upper row: Grasping a rectangular object. Since CM and MP joints have two degrees of freedom, the tip of the thumb and the tip of the index finger can be parallel to each other.
    Lower row: Grasping a circular object. The index finger is wrapped around the object to fit, while the thumb is not bent as much and acts like a wall.
    (b) The posture of the F3 hand when grasping the objects corresponding to (a). The combination of sliding parallel motion fingers and adaptive rotational motion fingers enables reproducing a grasping form similar to that of a human.
	}
	\label{fig:top_human_f3}
\end{figure}

\color{black}
One of the key factors that contribute to the dexterity of the human hand is the presence of the thumb. The thumb is positioned opposable to the other fingers, and acts as a fulcrum when the other fingers grasp something. The thumb does not just act as a passive fulcrum; we actively change the pose of our thumbs depending on the size and shape of the object to be grasped or the manipulation action we want to perform. As visualized in Fig.~\ref{fig:top_human_f3}, when we grasp a square object with our index finger and thumb, we alter the pose of our thumb to oppose our index finger by making use of two degrees of freedom in Carpometacarpal (CM) and Metacarpophalangeal (MP) joints of our thumb so that we can grasp the object in the most stable manner~\cite{kapanji_1980}. Additionally, when grasping a wide object, the tip of the thumb and the index finger can be positioned higher by bending these joints in the thumb. On the other hand, when grasping a narrow object, the tip of the thumb and the index finger can be positioned lower.
When we grasp a round object with the index finger and the thumb, the index finger wraps around the circumference of the object, while the thumb does not bend much and acts like a wall to determine the relative position between the object and our hand. Accordingly, the posture and the position of the tip of the thumb hardly change in this case (Fig.~\ref{fig:top_human_f3}~(a)).

In this work, we present a hand we call the ``F3 hand'' that can grasp both thin and large objects with simple and flexible control, giving the hand functions that take inspiration from the index finger and thumb of a human.
This hand is constructed after the movements of a human hand with a parallel motion finger and a rotational motion finger that have an adaptive function (Fig.~\ref{fig:top_human_f3}~(b)).

The contributions of this work is to propose a new robot hand inspired by the human index finger and thumb structure that combines a parallel motion finger and a rotational motion finger with an adaptive function.
This hand makes it possible to grasp a variety of objects spanning a wide range from tiny to large including thin and fragile objects with a single robot hand.

\section{RELATED WORK}
Numerous robot hands have been developed thus far. Parallel grippers with parallel finger motion and hands that grasp by rotating the fingers are the most common. In recent years, due to their versatility, there has been an increase in adoption of those with a structural adaptive function created with or without using soft materials.

\subsection{Parallel Grippers}
A variety of robot hands/grippers have been developed to meet the diverse application requirements. The most widely used type is a parallel gripper (e.g., \cite{schunk_EGI_website}). 
The use of parallel grippers offers a number of advantages; 1) it is easy to plan and execute the grasping of the objects, in particular, thin objects such as coins and washers because the fingertip height does not change during the grasping motion, 2) the hand architecture is simple and various actuators can be used as a driving source such as pneumatic actuator or electric motor because it is driven with only one degree of freedom.
In addition to the features of the parallel gripper, other types of grippers have been developed such as those that can compactly grasp large objects~\cite{kobayashiDesignDevelopmentCompactly2019} and those that can specifically grasp thin objects~\cite{yoshimi2012_hand, watanabe2021_hand}.
However, with parallel grasping, the end-effector is in contact with the object to be grasped on the surface, so if the end-effector is to be used for grasping flexible or deformable objects, it is necessary to devise some way of placing an elastic material on the contact surface.

\subsection{Rotational Finger Type Robot Hands}
Numerous robot hands have also been developed with fingers that rotate around a fixed part to perform grasping (e.g., \cite{zimmer_HRC05_website}). 
These hands tend to be more compact when compared to parallel grippers. However, it is not typically easy for them to grasp a thin object such as toothpicks, just like a parallel gripper because the height of the fingertip changes according to the degree of opening and closing of the finger.
For this reason, many practical hands that have rotational fingers inherit characteristics of parallel grippers to some extent (e.g., \cite{onrobot_Rg2_website}) 
The original HSR gripper, which achieves high grasping performance, is also an example of such a gripper~\cite{yamamoto2019development}.

\subsection{Adaptive Hands}
\begin{figure}[t]
	\centering
	\includegraphics[width=0.95\linewidth, draft=false]{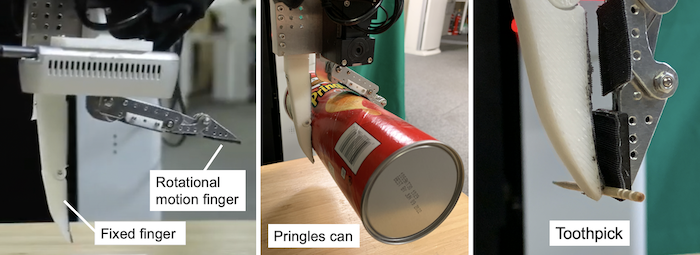}
	\caption{
	Overview of the F1 hand. A structure that combines a fixed finger with a rotational movable finger. The movable finger has an adaptive mechanism that allows the user to both hold the object as if it were wrapped around the object (center) and to pick up a toothpick (right) with the tip of the finger in a pinching motion.
	}
	\label{fig:f1_strucutre}
\end{figure}

Robot hands have been developed that attempt to grasp an object while maximizing contact over a larger area similar to the adaptive function of human fingers~\cite{stuart_cutkosky_marine_hand_2017, odhner_adaptive_precision_grasp_2012, yoon_flattouch_hand_2021}. The adaptive function of the hand allows grasping even with rough control, making it suitable for tasks that do not require precise placing, such as grasping and furniture manipulation.
In prior work, we developed a simple one-fingered hand called the F1 hand, which has an adaptive mechanism and drives only one movable finger, modelled after the index finger~\footnote{A promotional video of the hand with different attachments can be watched here: \url{https://www.youtube.com/watch?v=sN66DilliBs}}. The fixed finger is molded with resin and grasps the object in opposition to the movable finger. This fixed finger is not powered and can be easily replaced according to the application (Fig.~\ref{fig:f1_strucutre}).
Similar structures with a fixed finger on one side have also been developed in the past. For example, Pastor \etal{} developed a robot hand with a combination of fixed and working fingers with a large and stable surface for attaching tactile sensors to investigate in-hand manipulation~\cite{pastorUsing3dConvolutional2019}. They use the palpation motion of the working finger to recognize objects using a deep neural network. Ma \etal{} have also developed a structure that combines a fixed finger with a moving finger~\cite{maM2GripperExtending2016}. This F3 hand is built with two actuators on the moving finger to allow in-hand manipulation, so the approach is different from F1.
The F1 hand has a single actuator, which makes it simple and low-cost to build, but it also limits the tasks it can handle. In this case, the strategy is to expand the tasks to be handled by changing the fixed fingers.

The F1 hand can grasp objects that are difficult to grasp with normal fixed fingers by replacing them with fingers specially designed for the object. However, to avoid lengthening the working time, it is desirable not to make such an exchange.
Therefore, we decided to improve the function of the F1 hand by incorporating motion inspired by the functioning of the thumb as shown in Fig.~\ref{fig:top_human_f3}~(a) in addition to the adaptive function of index finger. Specifically, we developed a structure that combines a rotational motion finger that has an adaptive function and a parallel motion finger that can move in parallel like the thumb.

\section{HAND DESIGN}
\label{sec: hand_design}
\begin{figure}[t]
	\centering
	\includegraphics[width=0.95\linewidth, draft=false]{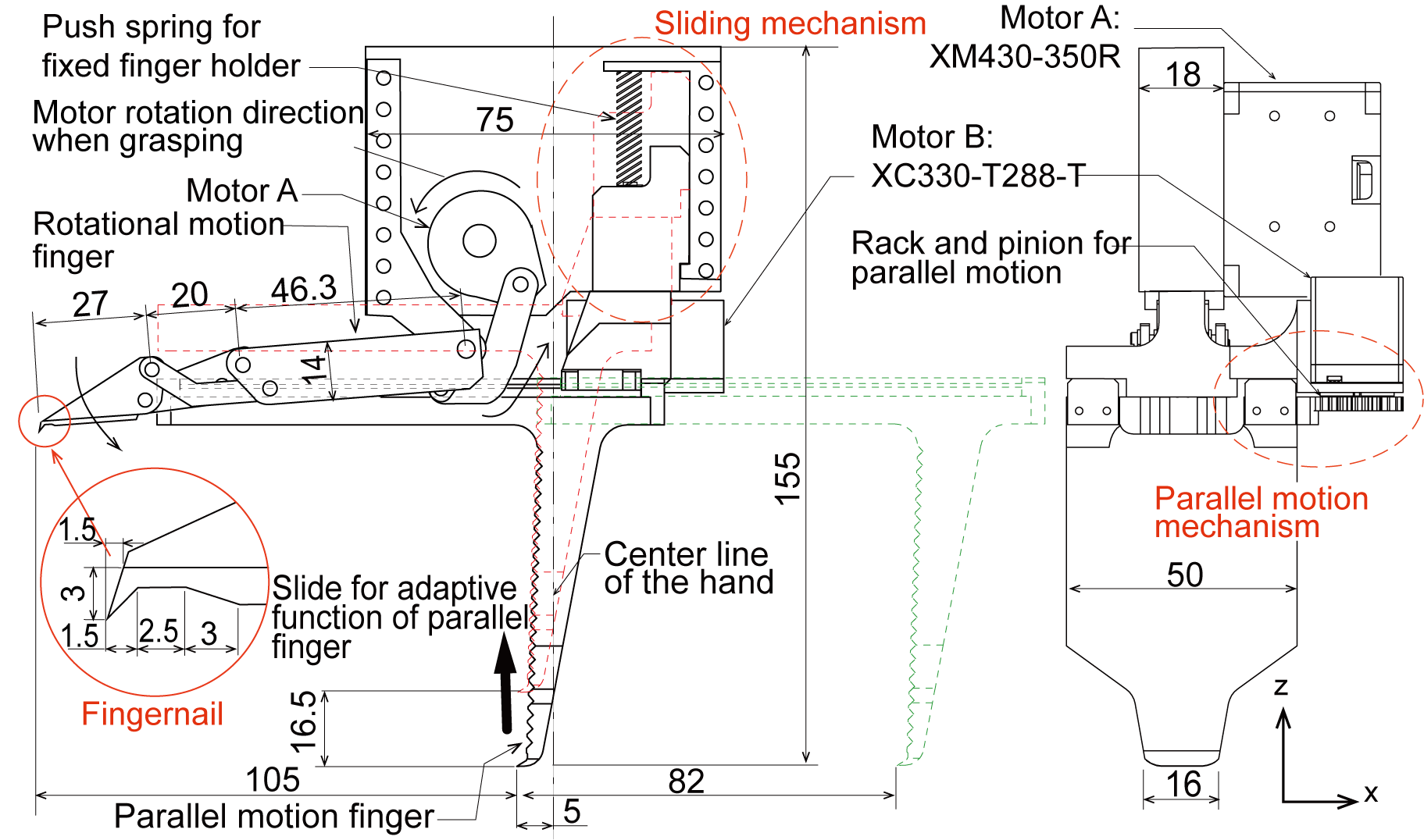}
	\caption{
	Structure of the F3 hand. Motors are placed in each of the rotational motion finger and parallel motion finger to drive them individually. The parallel motion finger is unitized and can be replaced according to the task. For example, if the application calls for holding a wider object, it can be replaced by a version with a longer rail.
	}
	\label{fig:f3_strucutre}
\end{figure}

\begin{figure}[t]
	\centering
	\includegraphics[width=0.95\linewidth]{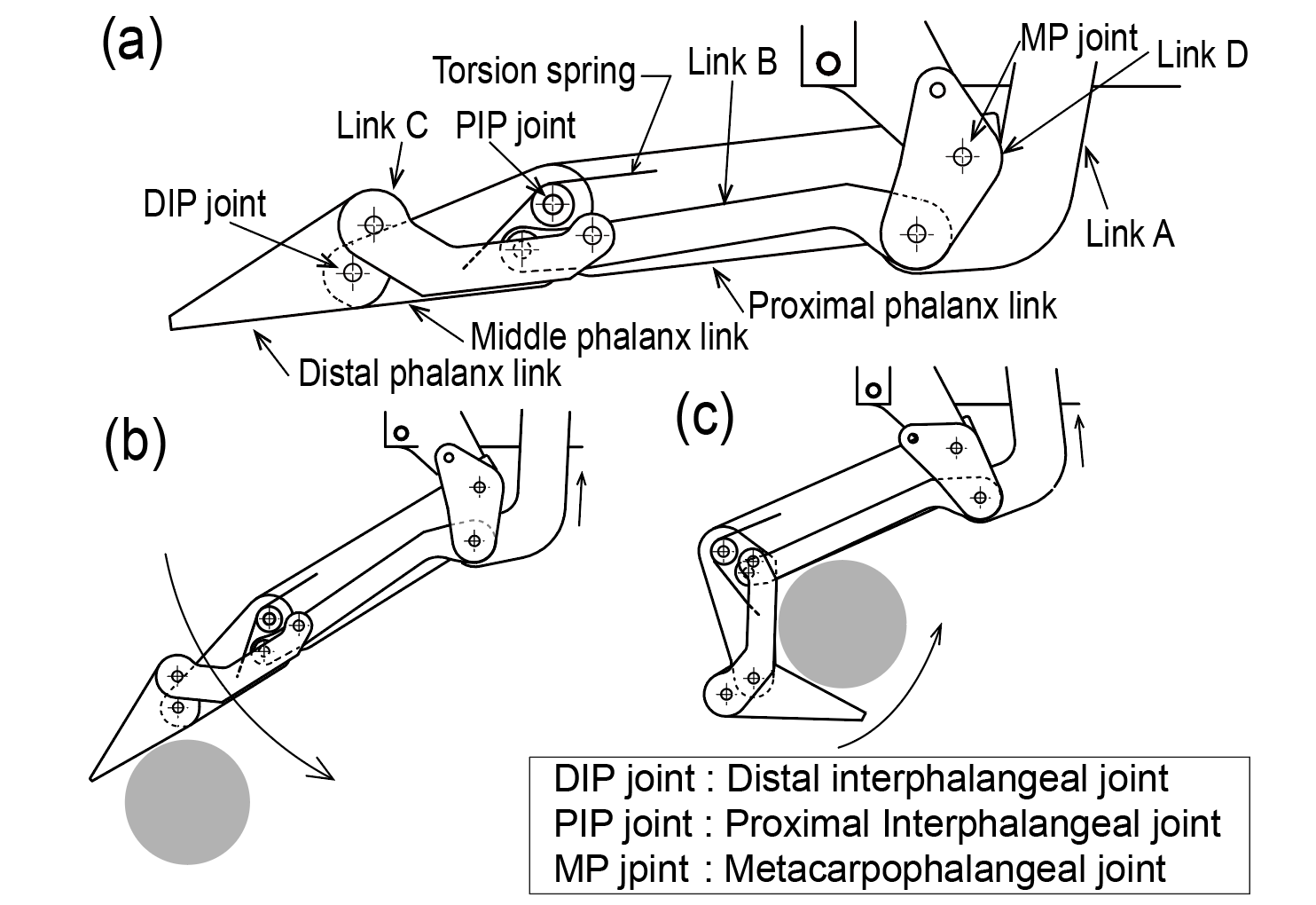}
	\caption{
	(a) Structure of the rotational motion finger.
	(b) Finger rotates while extended. When the fingertip hits an object, the contact force is increased while the finger is held extended.
	(c) When the middle part of the finger hits an object, it maximises contact area while equally distributing force exerted on the object as it wraps around it
	}
	\label{fig:f1fingerdesign}
\end{figure}

\subsection{Basic Design of F3 Hand}
Fig.~\ref{fig:f3_strucutre} shows the outline of the developed robot hand, F3 hand. It weighs $280~\mathrm{g}$ including the motor.
The finger mechanism with the adaptive function is shown in Fig.~\ref{fig:f1fingerdesign}. This adaptive finger mechanism is a space-saving version of finger the five-finger hand developed by one of the authors~\cite{fukaya2000_hand, fukaya2013_hand}. This mechanism is also used in F1 hands (Fig.~\ref{fig:f1_strucutre}).
When the distal phalanx link contacts the object, the grasping force is transmitted to where the rotational motion finger remains in extension (Fig.~\ref{fig:f1fingerdesign}~(b)).
When the proximal phalanx link contacts an object, link D moves independently and activates the middle phalanx link via link B, and link C (connected to the proximal phalanx link) pulls the distal phalanx link (Fig.~\ref{fig:f1fingerdesign}~(c)).
As a result, the movable finger moves in such a way that it wraps around the object, and the grasping motion automatically stops when all possible contact parts are achieved. 
Since the contact area is maximized, the contact force on the object is distributed across all the contact points, and the object can be stably grasped even with small motor torque.

This adaptive mechanism is also used in the movable fingers of our prior designed F1 hand.
However, the F1 hand has a weakness that it cannot grasp small objects such as the $6.4~\mathrm{mm}$ diameter washers included in the YCB object set~\cite{ycb_data_2015,calliYaleCMUBerkeleyDatasetRobotic2017a}, or large objects such as a mini soccer ball or a cracker box because one side of the finger is fixed.
On the other hand, F3 hand can solve these weaknesses by combining a rotational motion finger that has an adaptive mechanism as shown in Fig.~\ref{fig:f1fingerdesign} and a parallel motion finger that has a modularized mechanism that enables parallel motion with reference to the thumb. As shown in Fig.~\ref{fig:top_human_f3}~(a), the height of the tip of the thumb changes depending on the degree of opening and closing of the finger, so this parallel motion finger is designed to slide up and down from the fixed part.

By using the adaptive mechanism of the rotational motion finger and the mechanism that the parallel motion finger slides up and down, the hand absorbs the error to some extent even if the hand position or the opening/closing motion of the finger is rough. The Dynamixel XM430-W350 motor is used for the rotational motion finger side (Motor A), and the XC330-T288-T motor is used for the parallel motion finger side (Motor B). The parallel motion is operated by a rack and pinion mechanism attached to the side of the finger. Although the motor output of the parallel finger side is not so high, it is able to grasp heavy objects such as a frying pan due to the characteristic of the adaptive finger that automatically makes contact with the object over a larger area.

\subsection{Offset Function}

\begin{figure}[t]
	\centering
	\includegraphics[width=0.95\linewidth, draft=false]{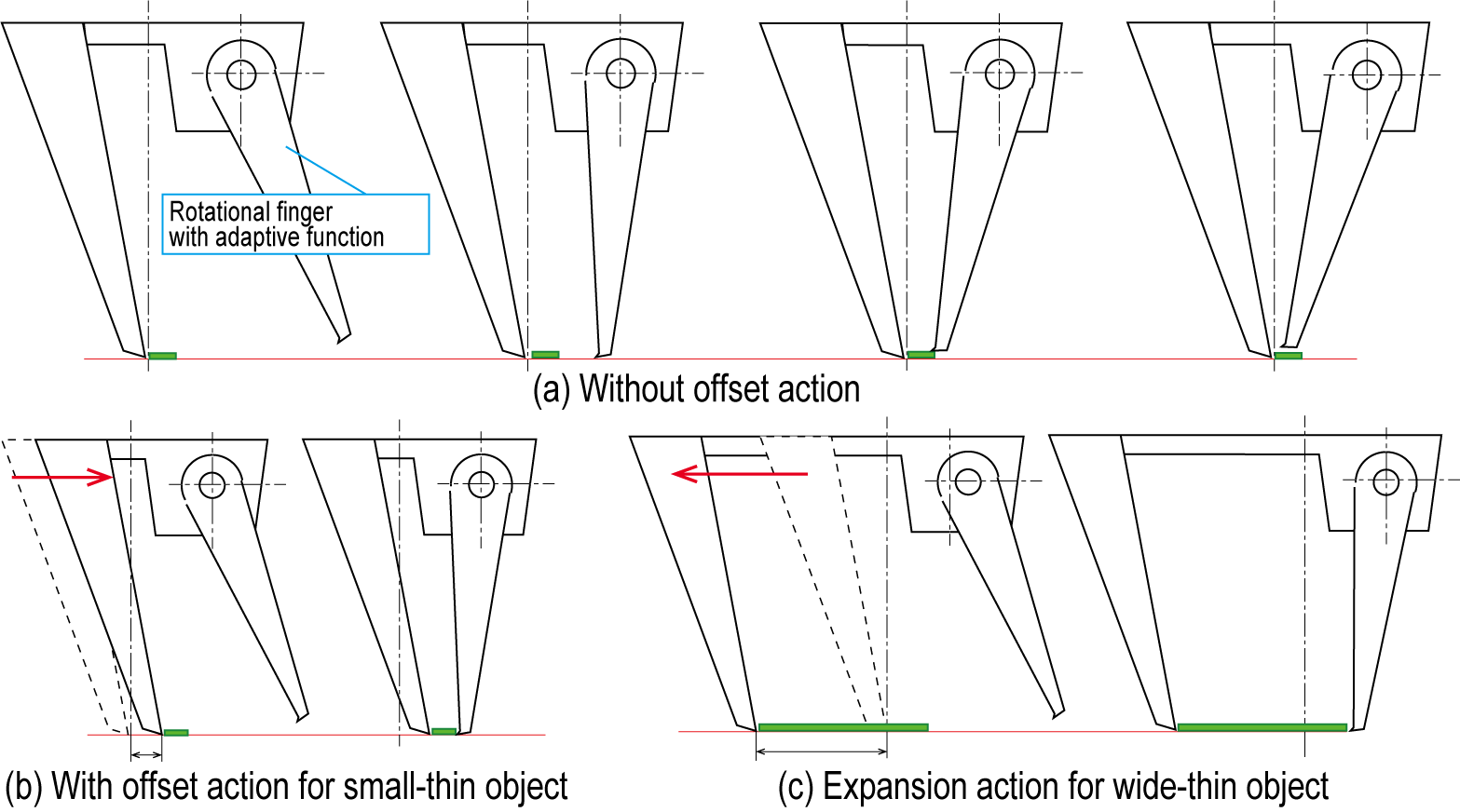}
	\caption{
	Improved grasping success rate by offset function. (a) The fingertip height changes slightly as the rotational motion finger turns to grasp. When grasping a thin object such as a small washer, this shift causes the grasp to fail. 
	(b) By moving the parallel motion finger offset from the hand center, the grasping can be done with the tip of the rotational motion finger at the lowest position. (c) For a thin and wide object like a credit card, the grasp success rate can be improved by widening the parallel motion fingers.
	}
	\label{fig:para_moion}
\end{figure}

Since the parallel motion finger can now be slid, we constructed it so that the tip of this finger can be offset from center of the hand to the rotational motion finger side. This makes it easier to achieve the posture that has the highest grasping success rate according to the shape and size of the object. For example, when grasping a small and thin object, in the case of a general rotating finger type hand or a hand like F1 hand, the structure that grasps the object by moving only the rotational motion finger changes the height of the tip of the finger and the fingertip is easily vertically displaced from the object to be grasped. For this reason, the F1 hand could not grasp extremely small and thin objects such as $6.4~\mathrm{mm}$ washers.
On the other hand, in F3 hand, the grasping success rate can be greatly improved by offsetting the parallel motion fingers so that the contact position of the rotational motion finger is at or around the vertical position when grasping the object (Fig.~\ref{fig:para_moion}).
This structure can also be used for grasping by moving the fixed finger while keeping the rotational motion finger in a fixed position. This is useful for scooping up and grasping small or thin objects, since the position of the fingertips and fingernails of the rotational motion finger remains almost unchanged.

\section{Experiment setup}
\label{sec:experiment_setup}

\subsection{Purpose of the Experiments}

\begin{figure}[t]
	\centering
	\includegraphics[width=0.95\linewidth, draft=false]{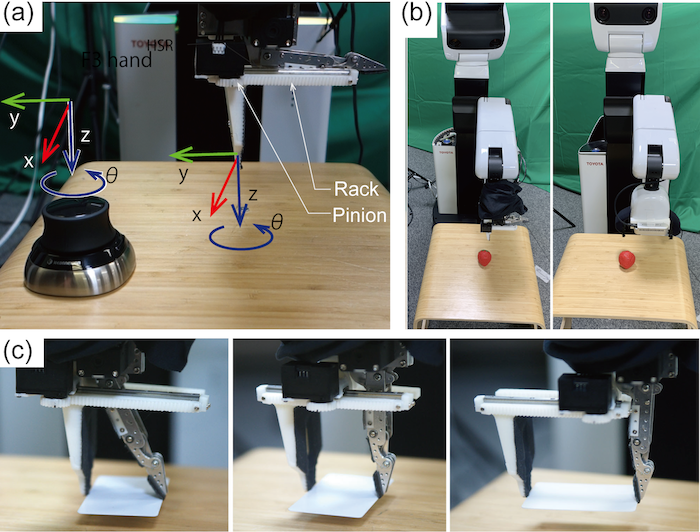}
	\caption{
	Experimental setup.
	(a) The 3D mouse interface. F3 hand/HSR gripper moves in the same direction as the 3D mouse knob is tilted.(b) The two HSRs used for experiments. It shows the teleoperation initial position. 
    (c) Various positional states of the parallel motion fingers. By using this functionality, we can choose amongst the various ways of grasping and holding the object and use the one that is most appropriate.
	}
	\label{fig:experimental_set}
\end{figure}

To confirm the characteristics of the F3 hand, these following points were validated.

    \begin{enumerate}
        \item Verification of the effect of using the structure and movement of the human hand as a reference
        \item Confirmation of the F3 hand's ability to grasp various objects used in daily life
        \item Reproducibility of human hand dexterity such as re-grasping and gentle grasping of soft objects.
    \end{enumerate}

For this purpose, we prepared an experimental system that allows teleoperation control of a mobile manipulator- the Toyota Human Support Robot (HSR)~\cite{yamamoto2019development}. 

The controller is an intuitive 3D mouse (3Dconnexion SpaceMouse Wireless), and each motor of the HSR and F3 hand is operated via ROS. The reason for using a mobile manipulator is to confirm the characteristics of F3 hand, which can easily perform grasping even with rough, imprecise positioning and grasping commands. In other words, it is to show that grasping can be performed in a situation closer to reality, instead of strictly controlling the robot arm numerically, as is commonly done in prior work on robot hands. Fig.~\ref{fig:experimental_set} shows the operation method.
The HSR, the hand motor, and the 3D mouse are connected to a PC running Ubuntu 18.04 and ROS melodic, and run through ROS.

\subsection{Overview of Experiments}
\label{sec:experimental_item}

To validate the basic characteristics of the F3 hand, various experiments were conducted on the following five themes.

\subsubsection{Confirmation of Reproducibility of Grasping Motion}

    In order to confirm whether grasping motion by parallel motion of the thumb like humans can be realized, grasping experiments were conducted on 7 types of washers in the YCB object set by sliding the parallel motion fingers.

\subsubsection{Grasping All Objects of YCB Object Set}
    To evaluate the versatility of the F3 hand, grasping experiments were conducted on all objects in the YCB object set. For most items, the parallel motion finger was set at the closest position to the hand center and the grasping was performed by moving only the rotational motion fingers. For the objects that were large in size such as a cracker box, large and heavy such as a wooden block, or petite and thin such as the smallest washer, the initial rotational motion finger was set fixed to point 90 degrees downward, and the parallel motion finger was moved for grasping behavior similar to that of a parallel gripper as shown in Fig.~\ref{fig:para_moion}~(c).

\subsubsection{Comparison of Grasping Performance with HSR Gripper}
    In order to confirm the level of grasping performance, we set the original HSR gripper and F1 hand as the baselines and compared the success rate and time required to grasp various kind objects. We picked 24 kinds of objects for grasping experiments, focusing on small, thin, complex, complicated shapes, and heavy objects. To clearly see the effect of the parallel motion fingers, the F1 hand is also equipped with a sliding mechanism for the fixed fingers shown in Fig.~\ref{fig:f3_strucutre}. Twenty grasping experiments were conducted for each object, and a total of more than 1,000 grasping experiments were conducted including the baseline experiment.
    
    Grasping experiments were limited to top grasping, which is used most frequently. For this reason, only the yaw angle $\theta$ and (x,y,z) directions were manipulated (see Fig.~\ref{fig:experimental_set}). For each object, 20-21 teleoperation experiments were performed by a skilled operator familiar with the 3D mouse interface and the HSR, with the goal of grasping as quickly as possible. To randomize the trials, the initial posture of the object and the grasping position of the robot were fixed, and the initial rotation of the wrist was set by sampling the rotation angle from a uniform distribution between $-45^\circ$ and $45^\circ$. This random motion requires the operator to always manipulate the robot to some degree to grasp the object. The robot was considered successful if it did not drop the object when it was automatically lifted to $10~\mathrm{cm}$. If the object fell slowly, e.g., by slipping, it was considered a failure.

\subsubsection{In-hand Manipulation}
    When a person grasps an object, he or she can change the posture and position of the object by dexterously moving the fingers in the hand. Such in-hand manipulation can be expected to improve positioning when placating an object, for example.
    Therefore, in order to perform such in-hand manipulation, we confirmed whether it is possible to manipulate the posture of an object after grasping it with a suction cup attached to the F3 hand.

\subsubsection{Soft Object Grasping}
    When gripping soft, easily deformable objects with a robot hand, holding the object in a point-contact manner may cause damage if the contact force is concentrated on a single point. The F3 hand is expected to have some usefulness in gripping such soft objects because it has a familiarization mechanism for the rotational motion fingers. Therefore, we verified the usefulness of two types of grasping methods for soft object grasping.

\section{EXPERIMENT}
\label{sec:experiments}
Grasping operation tests were conducted based on the experimental conditions presented in Section~\ref{sec:experiment_setup}. The results of each experimental item are shown below.

\subsection{Confirmation of Reproducibility of Grasping Motion}
\begin{figure}[t]
	\centering
	\includegraphics[width=0.95\linewidth, draft=false]{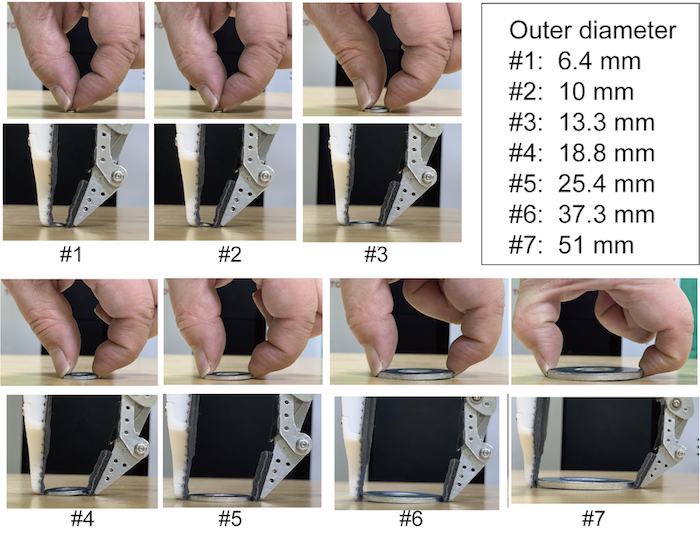}
	\caption{
	Comparison of grasping motions of seven different washers. By using the function shown in Fig.~\ref{fig:experimental_set}~(b), (c), it is possible to implement a grasping method with a high grasp success rate while maintaining a constant angle of the index finger as humans do.
	}
	\label{fig:fingetip_co}
\end{figure}

 Fig.~\ref{fig:fingetip_co} shows the result of the experiment. As shown in the figure, the rotational motion finger side, which corresponds to the index finger, bends the fingertip lightly like a human by touching the resting surface and blending in and the posture in which the fingernail makes contact with the washer side strongly is formed automatically. As a result, it is seen that all washers are successfully grasped. 
 Especially, thin objects are difficult to make contact in the appropriate position if only the rotational motion finger is moved, thereby changing the angle of the fingertip. In the case of small washers, this slight misalignment is enough to prevent gripping.
 On other hand, with F3 hand, the grasping can be done by sliding the parallel motion finger while keeping the posture of the rotary finger constant like the index finger. This indicates that the parallel motion of one of the fingers during the pinch grasp is useful for the fine grasping operation just like for a human.

\subsection{Grasping All Objects of YCB Object Set}

\begin{figure*}[t]
	\centering
	\includegraphics[width=0.95\linewidth]{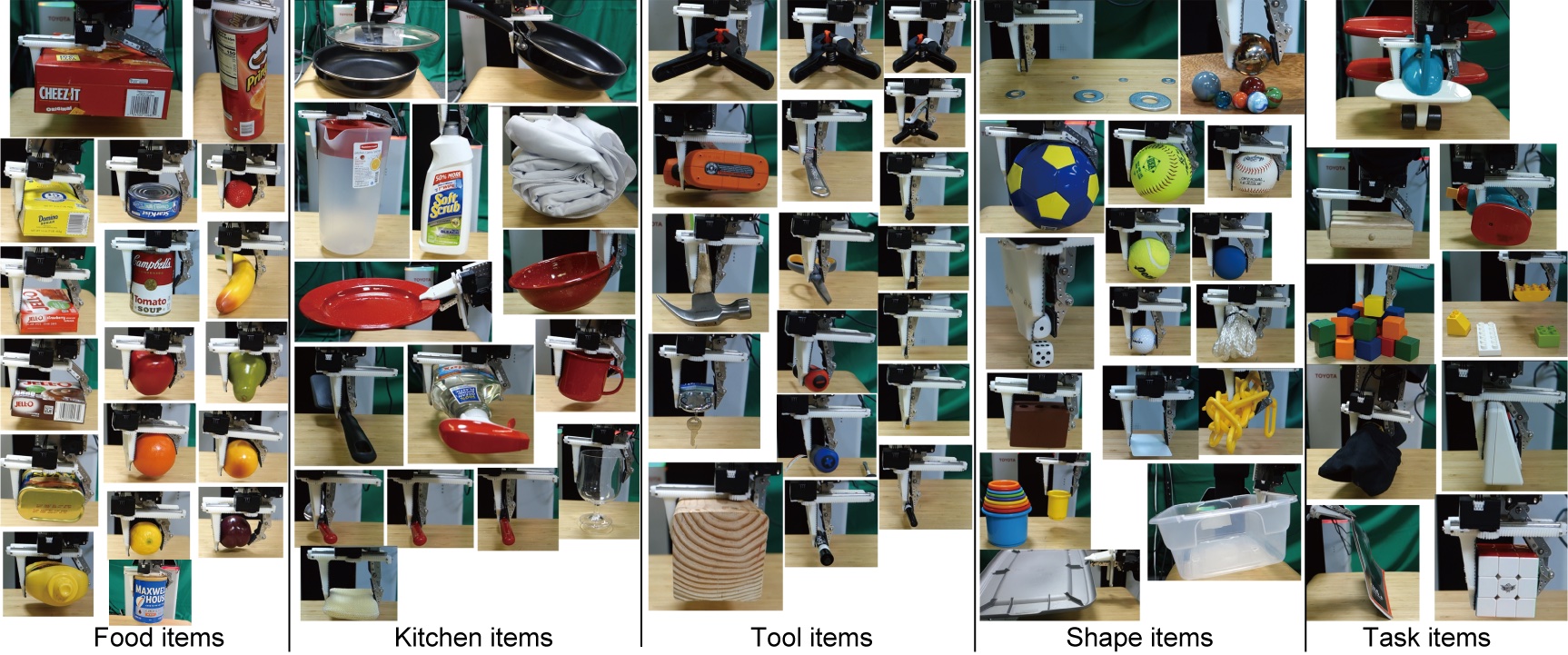}
	\vspace{-0.2cm}
	\caption{
		The objects in the YCB object set shown in the figure were grasped by teleoperation.
	}
	\label{fig:ycb_grasp}
\end{figure*}

As a result of the grasping experiment, we confirmed that the F3 hand can grasp all objects in the YCB object set ~\cite{calliYaleCMUBerkeleyDatasetRobotic2017a} shown in Fig~\ref{fig:ycb_grasp}.
T-shirts, tablecloths, and chains were grasped in the curled state. Since plates, lids of transparent boxes, and magazines could not be grasped with the top grasp, it was necessary to tilt the hand to the side and perform the grasping motion.

\subsection{Comparison of Grasping Performance with HSR Gripper}
\begin{figure}[t]
	\centering
	\includegraphics[width=0.95\linewidth,
	draft=false]{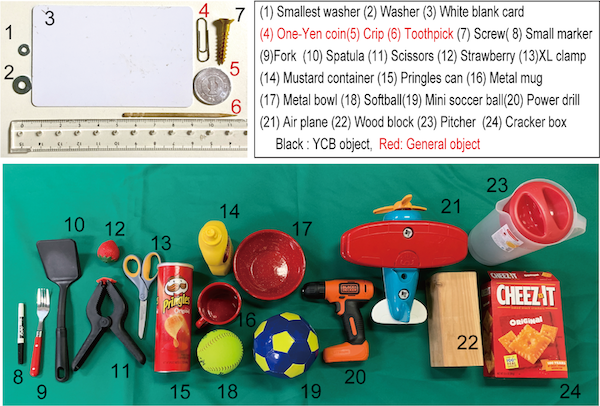}
	\vspace{-0.2cm}
	\caption{
		The objects used for experiments use a subset of the YCB objects~\cite{ycb_data_2015,calliYaleCMUBerkeleyDatasetRobotic2017a} with the addition of a Japanese one-Yen coin ,the toothpick and the paper clip.
		The pose of each object in the picture also indicates how these were placed on the table during top grasps.
		The numbers are used to identify each object.
	}
	\label{fig:objects}
\end{figure}

\subsubsection{Objects Used for Performance Experiments}
Many robot hands have been proposed to grasp small or large objects (e.g.~\cite{kobayashiDesignDevelopmentCompactly2019, yoshimi2012_hand, watanabe2021_hand}). Not only size, but also shape complexity and weight are points to be considered during grasping. 
Therefore, we decided to perform grasping experiments on the YCB object set~\cite{calliYaleCMUBerkeleyDatasetRobotic2017a} and several small objects to prove that our robot hand can grasp a wide range of shapes, sizes, and weights. Fig.~\ref{fig:objects} shows a subset of the 24 objects that we have chosen to grasp.
After confirming that the F3 hand is able to grasp all objects in the YCB dataset, we perform more thorough experiments by selecting 21 objects out of the YCB object set and added a small, thin Japanese one-Yen coin (diameter $20~\mathrm{mm}$ and $1~\mathrm{mm}$ thick), a small clip (length $27~\mathrm{mm}$, thickness $0.7~\mathrm{mm}$) and a toothpick (length $60~\mathrm{mm}$, diameter $2~\mathrm{mm}$). In addition, the smallest YCB washer (Outer diameter $6.4~\mathrm{mm}$, thickness $0.7~\mathrm{mm}$) is added to increase the difficulty.
Stable gripping of long items is a basic function required of robot hands. For this reason, we have selected elongated shapes for items \#8 through \#11. In grasping experiments, objects with complex shapes and curvatures are often selected for evaluation: the XL Clamp (\#13) and the Mastered container (\#14) were selected as objects with complex curvatures, and the pringles can (\#15) and metal mug (\#16) were selected as objects with many cups and similar sizes. Mini soccer ball (\#19) to Cracker box (\#24) were selected as representatives of large objects.

Our selection of objects covered the following ranges.
The heaviest object was the power drill ($895~\mathrm{g}$). 
The thinnest object were the smallest washer and the paperclip ($0.7~\mathrm{mm}$).
The smallest object was the washer ($\phi6.4~\mathrm{mm} \times 0.7~\mathrm{mm}$).
The maximum width object was the cracker box (bounding box approximately $230~\mathrm{mm} \times 160~\mathrm{mm} \times 60~\mathrm{mm}$).

\subsubsection{Results of Experiments}

\begin{figure}[t]
	\centering
	\includegraphics[width=0.95\linewidth, draft=false]{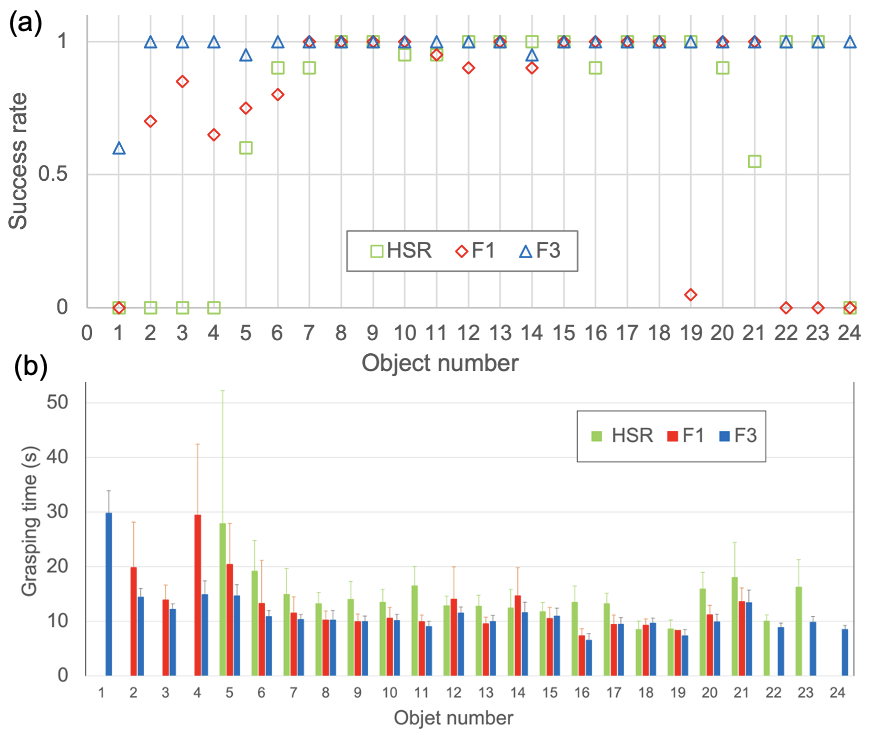}
	\caption{
	(a) Results of the teleoperation where each of the 24 objects were grasped between 20 to 21 times. 
	(b) Average time taken to grasp all the objects shown in Fig.~\ref{fig:objects} 24 times. 
	Original HSR gripper is simply denoted as `HSR'.
	Since the HSR's original gripper could not grasp objects \#1 -- \#4 and \#24 and F1 could not grasp objects \#1, \#22,\#23 and \#24, these values are not listed. On other hand, the HSR has suction cups, which could be used to grab items \#1 -- \#4. However, since this experiment was conducted to evaluate the gripper, the suction cups were not used.
	}
	\label{fig:success_rate_time}
\end{figure}

Fig~\ref{fig:success_rate_time}~(a) shows the grasp success rate. The aggregate success rates of the original HSR gripper and the F1 and F3 hands for all objects were 0.74, 0.73, and 0.98, respectively; original HSR gripper failed to grasp \#1 to \#4 for small objects and \#24 for a large object. The low success rate was mainly due to the large number of small items that were not grasped. 
F1 hand could not grasp the small item \#1, the large item \#19 only once, and items \#22 and above were never grasped successfully. This inability to grasp large objects had a significant impact on the lower success rate. In contrast, F3 was able to grasp all 24 objects by using parallel motion fingers.

Fig.~\ref{fig:success_rate_time}~(b) shows the results of the average time required for grasping.
Compared to HSR's original gripper, the F1 hand reduces gripping time by 10.5\% and the F3 hand by 20.5\%.
There are several reasons for this. The original HSR gripper has the fingertips move down as the fingers close, so the hand height and fingertip position must be fine-tuned while checking the height of the fingertips when gripping small or thin objects. In contrast, for F1 and F3 hands, the fingertips of the fixed fingers have obvious advantages when it comes to this. After bringing the fixed finger to the side of the object, the user only needs to execute the grasping motion, which simplifies the positioning.
Furthermore, F3 hand has a much higher success rate in grasping thin objects such as washers (\#2) than F1, and the time required for grasping is also much shorter. Normally, to increase the success rate of grasping thin objects, the heights of the left and right fingers should be pre-aligned. The fact that such adjustments can be easily implemented by sliding the parallel motion fingers is the reason why the F3 hand achieved higher results than the other two hands.

\subsection{In-hand Manipulation}
\label{sec:inhand}
\begin{figure}[t]
	\centering
	\includegraphics[width=0.95\linewidth, draft=false]{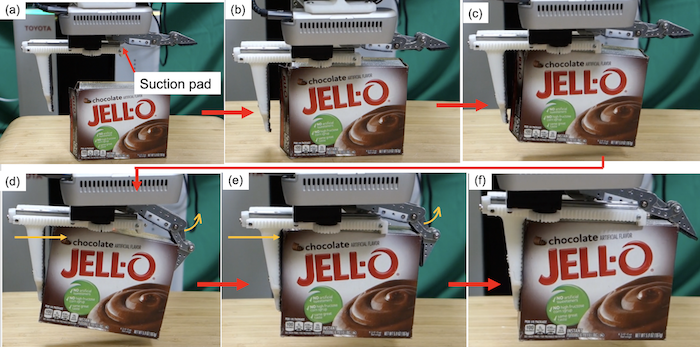}
	\caption{
	Example of changing posture and position after grasping an object.
    (a) Position of the added suction pad. (b), (c) Lifting an object by suction with the suction pad. (d) Grasping an object lightly with two fingers. (e), (f) Parallel motion fingers slid to fine-tune the position while keeping the object in the correct posture.
	}
	\label{fig:sucion_pic}
\end{figure}

By moving the parallel motion finger and the rotational motion finger separately, the object can be manipulated by the fingers. Fig.~\ref{fig:sucion_pic} shows a suction pad attached to the center of an F3 hand, ready to suck an object, and manipulate it when adsorbed.
Since the object can be handled like a mold that presses against the parallel motion fingers to adjust its posture, if the posture or position of the object is off during pickup, the suction cups can be used to hold the object while the rotational motion fingers support the sides of the object. It is also possible to move the object horizontally while in this state. This type of movement can also be carried out with the human hand, but not with as much horizontal movement as the F3 hand. This type of movement is unique to robots.

\subsection{Grasping of Fragile Objects}
\subsubsection{Soft object grasping motion utilizing parallel motion fingers}

\begin{figure}[t]
	\centering
	\includegraphics[width=0.95\linewidth, draft=false]{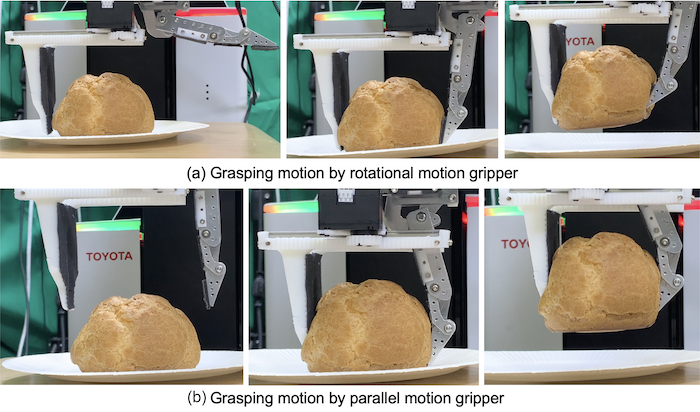}
	\caption{
    Grasping a cream puff. The deformation state of the object is changed by the different moving fingers. When the object is too soft, such as cream puffs, the grasping method using the parallel motion fingers is useful.
	}
	\label{fig:cream_puf}
\end{figure}

Fig.~\ref{fig:cream_puf}~(a) shows a cream puff grasped by moving the rotation motion finger. It can be seen that the cream puff is too soft, so no adaptive motion occurs when the rotational motion finger makes contact. As a result, it can be seen that the fingertips contact the cream puff without flexing, resulting in a point contact.

In contrast, when a grasping motion with parallel motion fingers was performed, which the F3 hand can perform, it was possible to form a grasping state with less load (Fig.~\ref{fig:cream_puf}~(b)).
As the parallel motion finger moves and presses the object against the rotation motion finger, this finger automatically changes its posture and adjusts so that it contacts the cream puff over a larger area. As a result, since a wider area is in contact with the object to be grasped, the object can be grasped with less deformation compared to a fingertip grasp. This characteristic is useful for grasping objects that are easily damaged, such as food and fruits.

\subsubsection{Fingertip Posture Suitable for Grasping Soft Objects}

\begin{figure}[t]
	\centering
	\includegraphics[width=0.95\linewidth, draft=false]{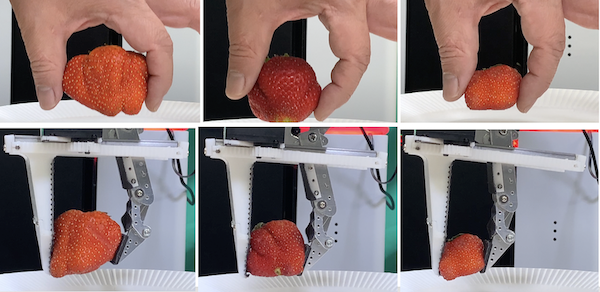}
	\caption{
    A human and F3 hand grasping strawberries of different sizes.
    It is important to adjust the opening and closing width of the fingertips according to size. This will make it easier for the fingertip of the index finger to make contact along the side of the strawberry. This way of holding the strawberry allows you to hold it over a larger area, thus making it less likely to damage the strawberry.
	}
	\label{fig:strawberry_grasp}
\end{figure}

When a human picks up something soft and of irregular size and shape such as a fruit, they adjust the posture of their thumb and index fingers in such a way that causes as little damage as possible. Therefore, we conducted an experiment to see if we could reproduce the same way of holding strawberries of different sizes after confirming the way a human grasps them.
 Fig.~\ref{fig:strawberry_grasp}~(a) shows the situation when a human grasps strawberries of three different sizes. 
As seen in Fig.~\ref{fig:fingetip_co}, the human can grasp the strawberries with the lowest possible force by moving each joint of the thumb, skillfully expanding the sensation between the thumb and the index finger according to the size of the strawberry, and lightly bending the index finger to contact the bottom surface of the strawberry.

Fig.~\ref{fig:strawberry_grasp}~(b) shows how the same strawberry is grasped by the F3 hand, using this grasping method as a reference. It can be seen that by sliding the parallel motion fingers, the holding method can be implemented such that the fingertips of the rotational motion fingers contact the bottom of the strawberry.

\section{CONCLUSION}

In this paper, we conducted experiments on the following items to confirm the features and versatility of the F3 hand we developed.
(1) Verification of the effect of using the structure of the human hand as a reference
(2) Confirmation of the F3 hand's ability to grasp various objects used in daily life
(3) Reproducibility of human hand dexterity, such as re-grasping the object to be grasped and the ability to safely grasp soft objects.

For (1), grasping experiments were conducted on seven different sizes of washers. By moving the parallel motion fingers according to the size of the washers, we were able to reproduce a posture similar to that of a person holding these washers (thumb up, index finger lightly bent) and confirmed that grasping was feasible.
For (2), grasping experiments were conducted on all YCB object sets in possession. We confirmed that all the other objects could be grasped with the top grasp with the exception of the plate, the lid of the transparent box, and the magazine which needed to be grasped with the hand pointing to the side. For comparison, we conducted grasping experiments with the original HSR gripper, the F1 hand, and the F3 hand on 24 objects of various sizes, and achieved grasping success rates of 0.74, 0.73, and 0.98 respectively. The average time required for grasping was reduced by 10.5\% for the F1 hand and 20.5\% for the F3 hand compared to the original HSR gripper.
For (3), we confirmed that by utilizing the parallel motion fingers, the rotary motion fingers can contact the appropriate position for grasping soft objects, such as cream puffs and strawberries, and that stable grasping can be performed. We also confirmed that the posture of the object after grasping can be adjusted by adding a suction cup and interlocking each finger after picking up the object.
We think the adaptive function of the finger contributes to the successful grasping of soft objects such as cream puffs and strawberries.
In this study, we performed all the experiments under the control of human teleoperation, but we think this adaptive function will be also important for the autonomous control in not well-known environment where the objects' poses and positions are not known in advance and the robot has to recognize the environment with its own sensors which provide, in general, imperfect and noisy observations.

F3 hand can increase the capability of the grasping owing to more diverse realizable grasping poses. 
On the other hand, unlike the HSR original gripper and F1 hand, which are driven by a single motor, the F3 hand is inferior in terms of cost and repairability because it has much complex structure and requires two motors.
It can increase the risk of breakdown compared to the F1 hand when a strong external force is applied to the robot hand. This is because parallel motion finger has gears which cannot absorb the force while
the rotational motion fingers can absorb the force due to the adaptive mechanism. Therefore, the choice of the hand should be considered depending on how it is used in the target application.

\section*{ACKNOWLEDGMENT}

The authors cordially express their gratitude to Mr. Shimpei Masuda, Dr. Tianyi Ko, Dr. Koji Terada and Dr. Kuniyuki Takahashi of Preferred Networks, Inc. and Mr. Koichi Ikeda, Mr. Hiroshi Bito and Dr. Hideki Kajima of Toyota Motor Corporation for their assistance.
	
\bibliographystyle{IEEEtran}
\bibliography{IEEEabrv, Fhands}

\begin{thebibliography}{10}
\providecommand{\url}[1]{#1}
\csname url@rmstyle\endcsname
\providecommand{\newblock}{\relax}
\providecommand{\bibinfo}[2]{#2}
\providecommand\BIBentrySTDinterwordspacing{\spaceskip=0pt\relax}
\providecommand\BIBentryALTinterwordstretchfactor{4}
\providecommand\BIBentryALTinterwordspacing{\spaceskip=\fontdimen2\font plus
\BIBentryALTinterwordstretchfactor\fontdimen3\font minus
  \fontdimen4\font\relax}
\providecommand\BIBforeignlanguage[2]{{%
\expandafter\ifx\csname l@#1\endcsname\relax
\typeout{** WARNING: IEEEtran.bst: No hyphenation pattern has been}%
\typeout{** loaded for the language `#1'. Using the pattern for}%
\typeout{** the default language instead.}%
\else
\language=\csname l@#1\endcsname
\fi
#2}}

\bibitem{kapanji_1980}
A.~I. Kapandji, ``The physiology of the joints, volume i, upper limb,''
  \emph{5th ed. Edinburgh: Churchill Livingstone}, 1982.

\bibitem{schunk_EGI_website}
Schunk, ``Egi series,'' Accessed on: march 7, 2022, [online], Available:
  https://schunk.com/jp\_en/gripping-systems/series/egi/.

\bibitem{kobayashiDesignDevelopmentCompactly2019}
A.~Kobayashi, J.~Kinugawa, S.~Arai, and K.~Kosuge, ``Design and {{Development}}
  of {{Compactly Folding Parallel Open-Close Gripper}} with {{Wide Stroke}},''
  in \emph{2019 {{IEEE}}/{{RSJ International Conference}} on {{Intelligent
  Robots}} and {{Systems}} ({{IROS}})}.\hskip 1em plus 0.5em minus 0.4em\relax
  {IEEE}, 2019, pp. 2408--2414.

\bibitem{yoshimi2012_hand}
T.~Yoshimi, N.~Iwata, M.~Mizukawa, and Y.~Ando, ``Picking up operation of thin
  objects by robot arm with two-fingered parallel soft gripper,'' \emph{IEEE
  International Workshop on Advanced Robotics and its Social Impacts(ARSO),},
  2012.

\bibitem{watanabe2021_hand}
T.~Watanabe, K.~Morino, Y.~Asama, S.~Nishitani, and R.~Toshima,
  ``Variable-grasping-mode gripper with different finger structures for
  grasping small-sized items,'' \emph{IEEE Robotics and Automation Letter},
  vol.~6, no.~3, pp. 5673--5680, 2021.

\bibitem{zimmer_HRC05_website}
Zimmer, ``Hrc-5 series,'' Accessed on: march 8, 2022, [online], Available:
  https://www.zimmer-group.com/en/technologies-components/components/handling-technology/grippers/hrc/collaborative/2-jaw-angular-grippers/hrc-05.

\bibitem{onrobot_Rg2_website}
OnRobot, ``Rg2 ^^e2^^80^^93 flexible 2 finger robot gripper with wide stroke,''
  Accessed on: march 6, 2022, [online], Available:
  https://onrobot.com/en/products/rg2-gripper.

\bibitem{yamamoto2019development}
T.~Yamamoto, K.~Terada, A.~Ochiai, F.~Saito, Y.~Asahara, and K.~Murase,
  ``Development of {{Human Support Robot}} as the research platform of a
  domestic mobile manipulator,'' \emph{ROBOMECH journal}, vol.~6, no.~1, p.~4,
  2019.

\bibitem{stuart_cutkosky_marine_hand_2017}
H.~Stuart, S.~Wang, O.~Khatib, and M.~R. Cutkosky, ``The ocean one hands: An
  adaptive design for robust marine manipulation,'' \emph{The International
  Journal of Robotics Research,}, vol.~36, no.~2, pp. 150--166, 2017.

\bibitem{odhner_adaptive_precision_grasp_2012}
L.~U. Odhner, R.~R. Ma, and A.~M. Dollar, ``Precision grasping and manipulation
  of small objects from flat surfaces using underactuated fingers,'' in
  \emph{2012 IEEE International Conference on Robotics and Automation}.\hskip
  1em plus 0.5em minus 0.4em\relax {IEEE}, 2012.

\bibitem{yoon_flattouch_hand_2021}
D.~Yoon, , and Y.~Choi, ``Analysis of fingertip force vector for pinch-lifting
  gripper with robust adaptation to environments,'' \emph{IEEE TRANSACTIONS ON
  ROBOTICS}, vol.~37, no.~4, pp. 1127--1143, 2021.

\bibitem{pastorUsing3dConvolutional2019}
F.~Pastor, J.~M. Gandarias, A.~J. {Garc{\'i}a-Cerezo}, and J.~M.
  {G{\'o}mez-de-Gabriel}, ``Using 3d convolutional neural networks for tactile
  object recognition with robotic palpation,'' \emph{Sensors}, vol.~19, no.~24,
  p. 5356, 2019.

\bibitem{maM2GripperExtending2016}
R.~R. Ma, A.~Spiers, and A.~M. Dollar, ``M2 gripper: {{Extending}} the
  dexterity of a simple, underactuated gripper,'' in \emph{Advances in
  Reconfigurable Mechanisms and Robots {{II}}}.\hskip 1em plus 0.5em minus
  0.4em\relax {Springer}, 2016, pp. 795--805.

\bibitem{fukaya2000_hand}
N.~Fukaya, T.~Asfour, R.~Dillmann, and S.~Toyama, ``Design of the
  {{TUAT}}/{{Karlsruhe}} humanoid hand,'' \emph{IEEE/RSJ International
  Conference on Intelligent Robots and Systems (IROS)}, pp. 1754--1759, 2000.

\bibitem{fukaya2013_hand}
N.~Fukaya, T.~Asfour, R.~Dillmann, and S.~Toyama., ``Development of a
  five-finger dexterous hand without feedback control: The
  {{TUAT}}/{{Karlsruhe}} humanoid hand,'' \emph{IEEE/RSJ International
  Conference on Intelligent Robots and Systems (IROS)}, pp. 4533--4540, 2013.

\bibitem{ycb_data_2015}
B.~Calli, A.~Singh, A.~Walsman, S.~Srinivasa, P.~Abbeel, and A.~M. Dollar,
  ``The ycb object and model set: Towards common benchmarks for manipulation
  research,'' \emph{2015 IEEE International Conference on Advanced Robotics
  (ICAR)}, pp. 510--517, 2015.

\bibitem{calliYaleCMUBerkeleyDatasetRobotic2017a}
B.~Calli, A.~Singh, J.~Bruce, A.~Walsman, K.~Konolige, S.~Srinivasa, P.~Abbeel,
  and A.~M. Dollar, ``Yale-{{CMU-Berkeley}} dataset for robotic manipulation
  research,'' \emph{The International Journal of Robotics Research}, vol.~36,
  no.~3, pp. 261--268, 2017.

\end{thebibliography}

\end{document}